\documentclass[
]{ceurart}

\usepackage{listings}
\usepackage{soul}
\usepackage{graphicx}  % for including images
\usepackage{caption}   % optional, for better caption formatting
\begin{document}

\copyrightyear{2025}
\copyrightclause{Copyright for this paper by its authors.
  Use permitted under Creative Commons License Attribution 4.0
  International (CC BY 4.0).}

\conference{Challenge and Workshop (BC9): Large Language Models for Clinical and Biomedical NLP, International Joint Conference on Artificial Intelligence (IJCAI), August 16--22, 2025, Montreal, Canada}

\title{Integrating Text and Time-Series into (Large) Language Models to Predict Medical Outcomes}
%\tnotetext[1]{You can use this document as the template for preparing your
%  publication. We recommend using the latest version of the ceurart style.}

\author[1,2]{Iyadh Ben Cheikh Larbi}[%
%email=iyadh.ben.cheikh.larbi@dfki.de,
]
%\cormark[1]
\address[1]{German Research Center for Artificial Intelligence (DFKI), Berlin, Germany} 
\address[2]{Technical University Berlin, Germany}

\author[1]{Ajay Madhavan Ravichandran}[%
%email=ajay\_madhavan.ravichandran@dfki.de,
]

\author[1]{Aljoscha Burchardt}[%
%email=aljoscha.burchardt@dfki.de,
]
%\address[1]{German Research Center for AI (DFKI), Germany}

\author[1]{Roland Roller}[%
email=roland.roller@dfki.de,
]
\cormark[1]
%\address[1]{German Research Center for AI (DFKI), Germany}

\cortext[1]{Corresponding author.}
%\fntext[1]{These authors contributed equally.}

\begin{abstract}
Large language models (LLMs) excel at text generation, but their ability to handle clinical classification tasks involving structured data, such as time series, remains underexplored. In this work, we adapt instruction-tuned LLMs using DSPy-based prompt optimization to process clinical notes and structured EHR inputs jointly. Our results show that this approach achieves performance on par with specialized multimodal systems while requiring less complexity and offering greater adaptability across tasks.

%While large language models (LLMs) excel at generative text-based tasks, their ability to classify patient data, including structured clinical information like time series, remains largely untapped. In this work, we explore how instruction-tuned LLMs, guided by DSPy prompt optimization, can be adapted to both clinical text and structured EHR data. Our results show that this lightweight, unified approach rivals specialized multimodal systems - offering a simpler, more flexible path toward clinical reasoning with LLMs.
\end{abstract} 


\maketitle

\section{Introduction}

Over the past decade, transformer-based models such as BERT have shaped the landscape of natural language processing (NLP), setting new standards for a wide range of tasks. More recently, large language models (LLMs) have emerged, challenging these `traditional' transformer models even on classical discriminative tasks despite not being explicitly designed for such purposes. However, the application of LLMs to more complex, multimodal data - such as electronic patient records - remains challenging. Patient records are multimodal in the sense that they are not only composed of text but also include structured numerical measurements that often add a temporal perspective.

In the past, approaches addressing this complexity typically involved elaborate architectures that combined different types of models, such as transformers with LSTMs, resulting in powerful but often difficult-to-maintain and little elegant solutions. In contrast, in this work, we explore how LLMs can be utilized to handle both modalities - unstructured text and structured time-series data - in a simple and unified manner. As a baseline, we take a traditional, smaller transformer model and show that the LLMs are not always the clear winner, especially when taking into account their resource needs, which may even be prohibitive in certain clinical setups.

To this end, we break down the problem into two parts: \textbf{(1)} evaluating how well LLMs perform on purely textual clinical NLP tasks (classification, inference, measuring similarity) compared to specialized clinical language models, and \textbf{(2)} investigating how LLMs can integrate temporal structured information, either directly or through textual descriptions, to improve clinical outcome prediction. Our research addresses the following questions: \textbf{(a)} Can instruction-tuned general-purpose LLMs match or outperform domain-specific clinical models on medical text-based tasks? \textbf{(b)} How can structured time-series data best be incorporated into LLM-based workflows to support clinical decision-making?

Through this work, we aim to better understand the capabilities and limitations of LLMs in healthcare applications that require reasoning across different types of data.

%\subsection{Questions to organizers}
%\hl{Double-blind or single blind? Appendix allowed? Acknowledgement\&Declaration on GenAI outside main paper?How many pages content? 4-6?}

\section{Related Work}

Past work on clinical tasks has used both unimodal models like LSTMs for time-series EHR data \cite{pmlr-v56-Choi16} and BERT variants for clinical notes \cite{naik-etal-2022-literature}. These were later extended to multimodal models using fusion techniques such as additive fusion \cite{khadanga-etal-2019-using}, gated fusion \cite{deznabi-etal-2021-predicting}, or cross-attention mechanisms \cite{zhang-etal-2022-pm2f2n, Qiao2019MNNMA}.

Recent studies explore LLMs for structured clinical data. \citeauthor{wu2024instruction} \cite{wu2024instruction} propose Llemr, which combines structured event embeddings from ClinicalBERT with LLMs, while we instead use DSPy prompting for more flexible fine-tuning. \citeauthor{liu2023fewshot} \cite{liu2023fewshot} demonstrate that LLMs can reason over physiological time-series data in few-shot settings. In contrast, our method explicitly adapts LLMs to structured clinical tasks via instruction-based fine-tuning. \citeauthor{Li2023FrozenLM} \cite{Li2023FrozenLM} focus on self-supervised alignment between ECGs and text using frozen LLMs, whereas we emphasize task-specific adaptation for multimodal clinical reasoning.

Unlike prior work, our approach aims to use instruction-tuned LLMs as a unified interface for both structured and unstructured clinical data, reducing the need for complex fusion architectures.

%Wu et al. (2024) \cite{wu2024instruction} introduce Llemr, a framework that uses ClinicalBERT to encode structured EHR events into embeddings, which are combined with textual data to fine-tune a large language model. Unlike their event-embedding approach, our method leverages DSPY prompting for fine-tuning, offering a more flexible and direct adaptation to downstream tasks.

%Liu et al. (2023) \cite{liu2023fewshot} explore the capability of large language models (LLMs) to perform few-shot learning on health-related tasks by grounding them in numerical time-series data like heart rate, step count, and sleep patterns. They show that with minimal examples, LLMs can reason over physiological and behavioral data for tasks such as cardiac signal analysis, activity recognition, calorie estimation, and mental health screening. Unlike our approach, which fine-tunes models using DSPy prompting with structured clinical instructions, their work focuses on minimal-tuning adaptation of LLMs to sensor-based health tasks.

%Li et al. (2023) \cite{Li2023FrozenLM} introduce METS, a method that uses a frozen language model to align ECG signals with clinical reports, enabling zero-shot classification without needing labeled data. Their focus is on leveraging self-supervised learning for multimodal signal-text alignment. In contrast, our approach fine-tunes models using DSPy prompting, which focuses on adapting large language models for clinical reasoning tasks based on structured EHR data, offering a more task-specific adaptation.
\section{Data \& Methods}

\subsection{Data}

We conduct experiments on four different medical datasets, three text-only tasks, and one task using EHR data, including text, as well as time-invariant and time-series data, as described below. All text-only tasks are solved as a multi-class problem using (micro/macro) F1. The multimodal task (mortality) is targeted as a binary classification but solved as a regression task, using AUROC and AUPRC for evaluation.

\paragraph{Smoking Status}
The Smoking Status dataset, created for the 2006 i2b2 Smoking Challenge \cite{2006smoking}, addresses multi-class classification based on discharge summaries. Given a summary, the task is to assign a patient to one of the following categories: \textit{Current smoker}, \textit{Past smoker}, \textit{Non-smoker}, \textit{Smoker} (unspecified), or \textit{Unknown}.

\paragraph{Natural Language Inference (MedNLI)}
MedNLI \cite{mednli} is a clinical natural language inference dataset derived from MIMIC-III. % \cite{mimic}. 
Given a premise and a hypothesis, the task is to classify their relationship as \textit{Entailment}, \textit{Contradiction}, or \textit{Neutral}.

\paragraph{Clinical Semantic Textual Similarity (ClinSTS)}
ClinSTS measures the semantic similarity between clinical sentence pairs \cite{clinsts}. To align this with the other tasks of this work, we follow previous papers that treat STS as a classification task, % (e.g. \cite{li2006sentence}, \cite{mihalcea2006corpus}, \cite{zhang2020enhanced}) 
and redefine the original continuous similarity scores (ranging from $0.0$ to $5.0$) as binary labels: pairs scoring above 3.0 are labeled as \textit{similar}, otherwise as \textit{dissimilar}.

\begin{comment} % If we want to use 4 classes for ClinSTS
We redefine the original continuous similarity scores (ranging from $0.0$ to $5.0$) as four labels that capture the semantic similarity to different degrees: 
\begin{itemize}
\item Pairs scoring between 0.0 and 1.5 are labeled as (1) dissimilar. 
\item Pairs scoring between 1.5 and 3.0 are labeled as (2) slightly similar.
\item Pairs scoring between 3.0 and 4.5 are labeled as (3) similar.
\item Pairs scoring between 4.5 and 5.0 are labeled as (4) identical. 
\end{itemize}
This can also demonstrate how the language models compare in ordinal classification.
\end{comment}

\paragraph{In-Hospital Mortality Prediction}
%The task, based on MIMIC-III data, % \cite{mimic}, 
%builds on semi-structured, longitudinal electronic patient records and targets the prediction of mortality of a patient after seeing the first 48 hours within the intensive care unit (ICU). We rely on the cohort of \hl{cite}, which includes 13 different features (mostly time-series, besides weight and height) and was extended by admission notes, as described in \cite{mortality_texts_preprocessing}, and preprocessed as in \cite{mortality_timeseries_preprocessing}, aggregating 48-hour features with standardized unit conversions and outlier handling. 

The task builds on the semi-structured electronic health records from MIMIC-III  %\cite{mimic} 
and aims to predict patient mortality in the intensive care unit using only the information recorded during the first 48 hours of hospitalization. It combines both textual and time-series data. For the textual component, we use the clinical admission notes as described in \cite{mortality_texts_preprocessing}. For the temporal component, we rely on the cohort of \cite{mortality_timeseries_preprocessing} that includes 13 features (mainly time-series data, besides weight and height). The 48-hour features are then aggregated along with standardized unit conversions and outlier handling.

\section{Models \& Techniques}

The baseline models for this work are BioClinical BERT \cite{bioclinical_bert} and Llama 3.1. %\cite{llama3_models}. 
These models differ significantly in architecture and intended use. BioClinical BERT is a popular encoder-only transformer model for clinical text classification tasks% (e.g., \hl{(citation)} \hl{(citation)})
, operating with an input context length of 512 tokens. In contrast, Llama 3.1 is a general-domain large language model capable of processing up to 128,000 tokens.

\subsection{BERT Fine-Tuning}
Encoder-only transformers like BERT are well-suited for discriminative tasks such as classification or named entity recognition. Fine-tuning typically involves extending the model with a linear layer as a classification head and training it on task-specific data to map embeddings to output labels. %\hl{(a bit more technical insights)}

\subsection{DSPy Prompt Optimization}
Prompting LLMs is highly sensitive to prompt formulation. DSPy \cite{dspy} provides a systematic framework for ``programming'' prompts rather than manually designing them. It uses algorithms like MIPROv2 %\cite{MIPROv2} 
to generate candidate instructions based on training data and task descriptions. DSPy then selects optimal instructions via Bayesian optimization according to a specified evaluation metric (e.g., classification accuracy). Example strategies include adding personas (e.g., ``you are a physician working in an ICU'') or emphasizing conciseness and clarity.

\subsection{LLM Instruction-Tuning}
Instruction-tuning is a supervised fine-tuning approach for LLMs, where the model learns to follow explicit task instructions. For classification tasks, the LLM is trained with prompt-output pairs: the prompt includes the task instructions and input examples, and the output corresponds to the correct class label. In this case, the model is trained using a causal language modeling objective, learning to predict the correct label based on the preceding context. However, following previous work making use of language model embeddings for classification, % (e.g., \cite{kant2018practical}, \cite{li2023label}, \cite{bert}), 
we introduce an additional objective, which consists of projecting Llama's last token output embedding through a simple linear layer to predict a class label. Similar to \cite{kant2018practical}, both the language modeling and the embedding classification are trained together, and the losses from both objectives are backpropagated through the model. % to teach the classification task. 
We opt for the ``Adapter-V2'' %\cite{adapterv2} 
parameter-efficient fine-tuning method to conduct the whole training while updating only 0.05\% of Llama's 8 billion parameters.

\section{Experiments}

\subsection{Setup}
%\hl{here some technical details about training}\\
Each experiment applies a model (Llama vs. BERT) and a corresponding method (prompting/instruction-tuning vs. fine-tuning) to achieve one of the tasks (Smoking Status, MedNLI, ClinSTS, In-Hospital Mortality). To obtain a fair evaluation, we ran each experiment three times and reported the median of the results. 
The experiments with BERT are all conducted on an RTX3090 (24GB) for five epochs. For the Llama model, a token limit of 2048 is sufficient to process all datasets. This allows the Llama experiments to be conducted on an RTXA6000 (48 GB) for five epochs. Since inference consumes less energy than fine-tuning, all evaluations and prompt optimization are performed on the RTX3090. The code base for the fine-tuning and evaluation of the models was adapted from the LitGPT %\cite{litgpt} 
implementation. 

%\paragraph{Text reduction} 

\subsubsection{Text Truncation}

Note that due to input length limitations, particularly for BERT models, the text data in the mortality task is truncated by discarding the excess tokens from the end of the sequence if the whole input exceeds the maximum context length. This is particularly important when adding time series data, as it makes input documents even longer. %\hl{it might be nice to see results without truncation on BERT}
Truncation might have an influence on the model performance, as possibly relevant textual information is lost. The average length of the text is 683.26 tokens. The average length of the numeric time series includes 309.09 tokens. Combining both sources will lead to text truncation in most cases, as both sources will likely exceed the maximum of 512 tokens (avg text length of truncated texts is 194.78).

%Note, due to input length limitations, particularly for BERT models, the time-series are aggregated into 8-hour intervals by computing mean values. If the combined input (text and time-series) still exceeds the maximum context length, the textual component is truncated (\hl{in which way?}) accordingly to ensure both modalities fit within model constraints. \hl{it might be nice to see results without truncation on BERT}

\subsubsection{Adding Time-Series}
\label{add_ts}

\paragraph{Numeric Integration} In the hospital mortality task, each patient record includes 13 different time-invariant \& time-series features collected over the first 48 hours of hospitalization. To reduce the number of values per feature, we represent each time-series feature within six mean scores covering the 48 hours. Then, all 13 additional features are appended to the text as a list of feature names, followed by a sequence of comma-separated values, as depicted in Figure \ref{fig:numeric_time_series}. % (see example in the Appendix in \ref{app:numberic_time_series}).  %In order to reduce the number of values per feature, we calculate the mean over each hour, thus each feature includes max 48 different values. All 13 additional features are appended to the text as a list of feature names, followed by sequence of comma-separated values (see example in the Appendix in \ref{app:numberic_time_series}). 

\begin{figure}[ht]
    \centering
    \tiny
    \fbox{\parbox{0.6\textwidth}{
        %\centering
        %\textbf{Example: Numeric Time-Series Data}\\
heart rate: 76.09, 78.75, 76.88, 69.75, 69.0, 69.0\\
respiratory rate: 19.19, 16.62, 16.62, 17.5, 16.0, 16.0\\
systolic blood pressure: 136.71, 129.94, 140.71, 144.56, 147.0, 147.0\\
diastolic blood pressure: 65.25, 56.06, 59.69, 55.81, 56.0, 56.0\\
mean blood pressure: 85.62, 76.5, 80.42, 78.5, 79.0, 79.0\\
oxygen saturation: 97.08, 96.38, 96.62, 96.38, 96.0, 96.0\\
temperature: 36.68, 36.56, 37.06, 37.0, 37.0, 37.0\\
glucose: 165.0, 127.0, 128.0, 128.0, 128.0, 128.0\\
Glasgow coma scale total: 11.0, 11.0, 11.0, 11.0, 11.0, 11.0\\
ph: 7.4, 7.4, 7.4, 7.4, 7.4, 7.4\\
fraction inspired oxygen: 0.21, 0.21, 0.21, 0.21, 0.21, 0.21\\
weight: 90.0\\
height: 170.0 
    }}
    \caption{Example Numeric Time-Series Data: 48 hours of values reduced to six mean values over time}
    \label{fig:numeric_time_series}
\end{figure}

%In the hospital mortality task, structured time-series data are incorporated directly into the input prompt. Each patient record includes 13 time-series features \hl{which ones?} , collected hourly over the first 48 hours of hospitalization. Each feature is formatted as a sequence of comma-separated values preceded by the feature name. \hl{show example of data with TS}

%\subsubsection{Time-Series Textual Interpretation}

\paragraph{Time-Series Description} In addition to inserting raw time-series values, we generated high-level textual interpretations of the clinical time-series data to enhance model understanding. A structured prompt was designed, outlining four key aspects: (1) overall stability of vital signs, (2) significant deviations from normal ranges, (3) notable trends and volatility, and (4) potential clinical concerns. This prompt, inspired by ChatGPT, was used to instruct Llama 3.1 to generate short (maximum five-sentence) summaries for each patient record. The interpretations avoid specific numerical values and instead focus on summarizing patterns and clinically relevant anomalies. An example prompt and the textual time-series description are presented in Figure \ref{fig:example_prompt} and Figure \ref{fig:time_series_description}, given the exemplary numeric time-series data, as shown in Figure \ref{fig:numeric_time_series}. %An example prompt and generated description are provided in Appendix~\ref{appendix:timeseries_description}.

\begin{figure}[ht]
    \centering
    \scriptsize
    \fbox{\parbox{0.97\textwidth}{
        %\centering
Analyze the following time series of aggregated clinical measurements from the first 48 hours of hospitalization. Provide a high-level interpretation (maximum 5 sentences, avoid specific numbers).

\begin{enumerate}
\item *\textbf{Overall Stability:}* Briefly state if vital signs are generally stable and within expected physiological ranges.
\item *\textbf{Deviations:}* Clearly identify any parameters consistently or significantly outside of typical normal ranges (mention if high or low).
\item *\textbf{Trends \& Volatility:}* Note any significant upward or downward trends, or if values show notable instability ('jumping').
\item *\textbf{Clinical Concern:}* Highlight any patterns or combinations of values that suggest potential clinical risk or deterioration.
\end{enumerate}

**[Insert \textit{Numeric Time-Series} Data Here]**

    }}
    \caption{ChatGPT-generated prompt to convert numeric-time series data into a time-series description.}
    \label{fig:example_prompt}
\end{figure}

\begin{figure}[ht]
    \centering
    \scriptsize
    \fbox{\parbox{0.97\textwidth}{
        %\centering
 The patient's heart rate is relatively stable, but there are some minor fluctuations. The respiratory rate is also consistent, indicating no immediate concerns. However, the systolic blood pressure shows a slight increase over time, which may be worth monitoring. The oxygen saturation levels are within normal range, and the temperature is slightly elevated.

    }}
    \caption{Time-series description: Response given the prompt above and the exemplary numeric time-series}
    \label{fig:time_series_description}
\end{figure}

\subsection{Results}

%\subsubsection{Text-only Tasks}

\paragraph{Text-only Tasks} We first evaluate performance on tasks that rely solely on textual input: MedNLI, Smoking Status, and ClinSTS. As presented in Table \ref{tab:results_large}, across all three tasks, the instruction-tuned Llama model consistently outperforms both the DSPy Llama (using optimized zero-shot prompting) %\hl{(zero/few-shot)}
and BioClinicalBERT. Specifically, instruction-tuned Llama achieves the highest macro and micro scores for MedNLI (0.89), Smoking Status (0.63/0.80), and ClinSTS (0.92/0.94). BioClinicalBERT shows competitive but lower performance, particularly on MedNLI and ClinSTS, while DSPy Llama underperforms across all text-only tasks, reflecting the advantage of explicit supervised fine-tuning in clinical settings.

\begin{table}[htbp!]
\scriptsize
    \begin{tabular}{c|l|cc|cc|cc|cc}
    \hline
     &  & \multicolumn{2}{c}{\textbf{MedNLI}} & \multicolumn{2}{c}{\textbf{Smoking Status}} & \multicolumn{2}{c}{\textbf{ClinSTS}}  & \multicolumn{2}{c}{\textbf{Mortality}}  \\ \hline
     & Setup   & Macro & Micro & Macro & Micro & Macro & Micro & AUROC & AUPRC  \\ \hline
    \multirow{3}{*}{\rotatebox[origin=c]{90}{text}} 
     & DSPy Llama 3.1 & 0.67 & 0.67 & 0.50 & 0.72 & 0.79 & 0.84 & 51.37 & 8.99 \\
     & BioClinical BERT & 0.81 & 0.81 & 0.53 & 0.76 & 0.86 & 0.89 & 83.45 & 39.22 \\
     & Inst. Tuned Llama 3.1 & \textbf{0.89} & \textbf{0.89} & \textbf{0.63} & \textbf{0.80}  & \textbf{0.92} & \textbf{0.94} & \textbf{85.58} & \textbf{44.78} \\ \hline
    \multirow{7}{*}{\rotatebox[origin=c]{90}{text+TS}} 
     & DSPy Llama 3.1 & & & &  & &  & 59.84 & 11.95 \\
     & BioClinical BERT & & & &  &  & & 89.52 & \textbf{64.03}  \\ 
     %& BioClinical BERT (TS-GCS) & & & &  &  & & 89.34 & 61.12 \\ 
     %& Inst. Tuned Llama 3.1 (48TS) & & & &  & &  & \textbf{91.15} & 62.51 \\ 
     & Inst. Tuned Llama 3.1 & & & &  & &  & \textbf{90.82} & 60.14 \\ 
     & BioClinical BERT (descr) & & & &  & &  & 86.64 & 52.34 \\ 
     & Inst. Tuned Llama 3.1 (descr) & & & &  & &  & 88.91 & 56.73 \\ \hline
    %\multirow{2}{*}{\rotatebox[origin=c]{90}{TS}} 
     %& BioClinical BERT & & & &  & &  & 88.72 & 62.89 \\ 
     %& Inst. Tuned Llama 3.1 & & & &  & &  & 88.16 & 62.46 \\ \hline
    \end{tabular}
    \caption{Macro/Micro F1 Scores for medical text tasks (MedNLI, Smoking Status and ClinSTS) and AUROC and AUPRC results for mortality prediction. Experiments are conducted only with text and text+time series (TS).}
    \label{tab:results_large}
\end{table}

%\subsubsection{Mortality Prediction (with Time Series)}

\paragraph{Mortality Prediction (with Time Series)} In the mortality prediction task, where models can leverage structured time-series data, a clear performance trend emerges (see right side of Table 1). When using only text, the instruction-tuned Llama 3.1 model outperforms both DSPy Llama and the fine-tuned BioClinical BERT, indicating strong overall discrimination and precision-recall balance. The addition of time-series data leads to consistent improvements across models, confirming the utility of multimodal input. In particular, numeric integration of time-series features proves more effective than descriptive text integration. Interestingly, while the instruction-tuned Llama achieves the highest AUROC, BioClinical BERT with numeric time-series input obtains the best AUPRC score - highlighting a trade-off between overall classification performance and performance on the positive class. This demonstrates the added value of structured temporal data and underscores the importance of evaluating models on multiple complementary metrics. %Although prior work has observed similar trends, this is the first time such findings are shown for LLM-based models in this clinical context.
%In the mortality prediction task, where models can leverage structured time-series data, a similar trend emerges (see right side of Table \ref{tab:results_large}). First, using only text, the instruction-tuned Llama outperforms both DSPy Llama, as well as the fine-tuned BioClinical BERT. When adding time-series data, we can see the same pattern: the fine-tunes Llama outperforms BioClinical BERT, whereas the integration of numeric information seems to be the more efficient way to integrate time-series, in contrast to the descriptive integration of time-series data. However, in all cases, the benefit of additional using time-series data is clearly visible, which is in line with other related work but has not been shown in this context for LLMs. Finally, surprisingly, BioClinical BERT with numeric integration outperforms all approaches on AUPRC.

%When text data is combined with raw time series (TS), BioClinicalBERT reaches a ROC-AUC of 89.52 and a PRC of 64.03, outperforming its text-only version by a significant margin. Instruction-tuned Llama 3.1 achieves an even higher ROC-AUC of 90.82 with time series, although its PRC (60.14) slightly lags behind BioClinicalBERT. When instead using textual descriptions of time-series data (instead of raw values), performance slightly drops for both models but remains above their text-only baselines (e.g., BioClinicalBERT achieves ROC-AUC 86.64 with descriptions vs. 83.45 with text only). These results indicate that access to structured or summarized time-series information substantially boosts mortality prediction performance.

\subsection{Further Analysis}
%\hl{some analysis: only TS, ablation, all 48 values, then electricity}

In the following, we explore the benefit of integrating time-series data in text a bit more in detail. First of all, in the case of instruction-tuned Llama, we can increase the number of input tokens compared to BERT-based models, thus using a model with 48 values per feature (averaged per hour) instead of 6 values as tested above (see Sec \ref{add_ts}), the performance further increases to AUROC 91.15 and AUPRC 62.51, which is still below the best AUPRC of the BioClinical BERT. Moreover, we were generally surprised that language models - large and normal - are capable of dealing with numeric information in text, so we fine-tuned both models solely on TS and achieved an AUROC of 88.72 (88.16) and AUPRC 62.89 (62.46) for BioClinical BERT (Llama) - both providing very similar and high scores, better than using only text. It is not surprising that time series might be more valuable for this task compared to text, but it is that both models can deal with numeric data so well. Removing, for instance, the feature Glasgow Coma Scale (GSC) leads to a drop of approx 3 points in AUPRC for BioClinical BERT (text+TS), although in 80\% of the cases, GSC is not mentioned. %\footnote{If GSC is below 8, the patient is in a serious condition, which indicates its importance and is probably learned by the different models.}. 

\begin{table}[htbp!]
    %\centering
    \tiny
    \begin{tabular}{l|cc|cc|cc|cc}
    \hline
      & \multicolumn{2}{c}{MedNLI} & \multicolumn{2}{c}{Smoking Status} & \multicolumn{2}{c}{ClinSTS}  & \multicolumn{2}{c}{Mortality}  \\ \hline
       & Energy & Time & Energy & Time & Energy & Time & Energy & Time  \\ 
    Llama Inst. tuning* & 375.54 & 1443.67 & 11.17 & 46.35 & 61.35 & 232.35 & 2106.93 & 7472.07 \\
    BERT Fine-tuning* & 8.17 & 44.86 & 0.35 & 3.39 & 1.69 & 8.43 & 43.77 & 166.44 \\
    Llama inference** & 3.47 & 11.18 & 14.52 & 42.32 & 3.66 & 11.77 & 13.25 & 38.65 \\ 
    Bert inference** & 0.14 & 1.26 & 0.18 & 1.66 & 0.11 & 0.99 & 0.15 & 1.27 \\
    Prompt optim. & 219.79 & 670.37 & 707.19 & 2147.71 & 204.7 & 640.54 & 1273.22 & 4512.63  \\  \hline
    \end{tabular}
    \caption{Time (in seconds) and Energy Usage (in kJ): *Instruction/Fine-tuning energy and time for 1 epoch; **Inference energy and time for 100 samples; > time and energy consumed depends not only on GPU and model used but also on the size of the dataset and the length of the prompts}
    \label{tab:energy_time}
\end{table}

Finally, in Table \ref{tab:energy_time}, we compare the energy consumption and time of fine-tuning and inferring one of our models. This is particularly relevant if time plays a crucial aspect and limited hardware capabilities are available. %

%\paragraph{Input Length BioClinical BERT} While for Llama the input length does not play an important role (it is always within the context length), for BioClinical BERT, it does, as it accepts only up to 512 input tokens. All tokens above are cut off at the end. This has an influence on the model, particularly when not truncated, as the average length of texts is 683.26 tokens, whereas only 194.78 on average with truncated texts. The average length of the numeric time-series includes 309.09 tokens. In this way, in most cases, the truncated text, in combination with the time-series, is within the legit context window.

\section{Discussion}

Our experiments reveal two key insights. First, in purely text-based tasks (MedNLI, Smoking Status, ClinSTS), the instruction-tuned Llama 3.1 outperforms both a domain-specific model (BioClinicalBERT) and a general LLM without fine-tuning (DSPy Llama 3.1). This underscores the value of supervised instruction tuning, even for large, general-purpose models applied to specialized medical domains.

Second, incorporating time-series data substantially improves performance in the downstream clinical prediction task (mortality). Direct access to raw numerical measurements yields the highest scores, but even textual descriptions of temporal trends lead to noticeable gains over using text alone. This suggests that well-crafted summaries of time-series patterns can serve as effective substitutes when numeric data is unavailable or difficult to process.

Together, these results confirm our hypothesis: enriching clinical language models with structured temporal information - raw or summarized - enhances predictive performance. While Llama 3.1 shows clear advantages in text-only tasks, its large size may be a limiting factor in resource-constrained settings. In the multimodal mortality task, BioClinicalBERT offers a strong trade-off: it trails Llama slightly in AUROC but outperforms it in AUPRC. Given these considerations, we would favor the smaller, specialized model when hardware or deployment constraints are present.

\section{Conclusion}

Our findings show that instruction-tuned general-purpose language models, such as Llama 3.1, can outperform domain-specific transformers like BioClinicalBERT on text-only clinical tasks - highlighting the impact of supervised instruction tuning in specialized domains. In clinical prediction tasks such as mortality, integrating structured time-series data - either as raw numeric input or as textual summaries - consistently improves model performance. Notably, while Llama achieves the highest AUROC scores, BioClinicalBERT performs best on AUPRC, reflecting a trade-off between overall discrimination and sensitivity to the positive class.

These results emphasize the value of combining large language models with structured multimodal inputs and demonstrate that even smaller, domain-specific models remain competitive, especially when hardware efficiency or deployment constraints matter. Furthermore, the ability of LLMs to process numerical data embedded in text opens up promising avenues for low-resource or interoperable medical AI systems, where direct access to structured data may be limited. Overall, both fine-tuning and multimodal integration prove essential for developing robust, practical AI tools in healthcare.

%Our study demonstrates that instruction-tuned general-purpose language models can outperform traditional clinical transformers on text-based clinical tasks. Moreover, augmenting these models with structured temporal information — either as raw time series or as textual summaries — significantly enhances their performance in clinical prediction tasks. These findings highlight the value of both fine-tuning and multimodal data integration in building more effective AI systems for healthcare and suggest that even lightweight textual representations of time series can meaningfully contribute when direct numerical data integration is challenging.

%\section{Acknowledgments}

\begin{acknowledgments}
The project has received funding from the Federal Ministry of Research, Technology and Space (BMFTR) through the projects KIBATIN (16SV9040), PRIMA-AI (01GP2202C) and the Federal Joint Committee of Germany (Gemeinsamer Bundesausschuss) as
part of the project SmartNTX (01NVF21116).
\end{acknowledgments}

%% The declaration on generative AI comes in effect
%% in January 2025. See also
%% https://ceur-ws.org/GenAI/Policy.html
\section*{Declaration on Generative AI}
The authors used GPT-4 and Grammarly to improve the overall quality of the text (removing typos and re-writing unclear sentences). After using these tools, the authors reviewed and edited the content and take full responsibility for the content. 

%%
%% Define the bibliography file to be used
%\bibliography{sample-ceur}

\end{document}